# A technology-oriented mapping of the language and translation industry – Analysing stakeholder values and their potential implication for translation pedagogy

María Isabel Rivas Ginel, Université catholique de Louvain, isabel.rivasginel@uclouvain.be

Janiça Hackenbuchner, Ghent University, janica.hackenbuchner@ugent.be

Alina Secară, University of Vienna, alina.secara@univie.ac.at

Ralph Krüger, TH Köln – University of Applied Sciences, ralph.krueger@th-koeln.de

Caroline Rossi, Université Grenoble Alpes, caroline.rossi@univ-grenoble-alpes.fr

## Abstract

This paper examines how value is constructed and negotiated in today's increasingly automated language and translation industry. Drawing on interview data from twenty-nine industry stakeholders collected within the LT-LiDER project, the study maps their views to the four categories of *human value*, *technological value*, *efficiency*, and *adaptability*. Using Chesterman's framework of translation ethics and associated values as an analytical framework, the paper shows that efficiency-oriented technological values aligned with the ethics of service have become baseline expectations in automated production environments, where speed, scalability, and deliverability dominate evaluation criteria. At the same time, human value is repositioned, associated primarily with aspects such as expertise, oversight, accountability, and contextual judgment in technology-assisted workflows. Adaptability is constructed as a core professional requirement, reflecting expectations that translators continuously adjust their skills, roles, and identities in response to evolving tools and organisational demands. Overall, the picture of an interdependent configuration emerges, in which technological efficiency enables human multilingual communication. To conclude, the paper discusses the potential implications of our technology-oriented and value-focused analysis of the contemporary language and translation industry for translator education.

**Keywords:** Language and translation industry; automation; translation technology; stakeholder values; efficiency; adaptability

## Introduction: On the evolving notion of "value" in the AI-automated language and translation industry

In recent years, the language and translation (L&T) industry has undergone profound transformations driven by technological, as well as associated economic and discursive shifts. Chief among these has been the accelerating automation of translation through high-performing artificial intelligence (AI) technologies, first in the form of neural machine translation (NMT) and more recently through large language models (LLMs), a category of general-purpose AI (GPAI) (cf., e.g., ELIS Research 2026). In addition to altering professional practices and workflows, these developments have also affected the value of translation and translators as perceived by different industry actors. Against this backdrop, this paper investigates the kinds of values highlighted in a series of interviews by various stakeholders occupying different professional roles in the contemporary L&T industry and examines the implications of these value configurations for translation pedagogy.

Merriam-Webster defines *value* as the "relative worth, utility, or importance"[1] of a tangible or intangible phenomenon. This definition highlights that value is usually not an absolute measure but rather established in relation to one or more frames of reference. In the case of complex and multidimensional social phenomena such as translation, it could be claimed that these frames of reference are culturally situated, emerge from and are shaped by particular cultural environments. Concerning translation, Prunč's (1997) notion of *translation culture* ('Translationskultur') captures this dynamic well by referring to "the self-regulating translation-related subsystem in a culture that has developed historically out of initially unstructured fields of translation practice [and that] can be differentiated into different subcultures" (Risku & Windhager, 2013, p. 34). The international L&T industry may be understood as such a translational sub- or diaculture, further decomposable into multiple microcultures, including institutional settings such as the European Commission's Directorate-General for Translation or commercial language service providers like RWS or Lionbridge. These sub-, dia-, and microcultures form a complex relationship with each other, as well as with relevant paracultures that configure and prioritise the frames of reference relative to which the value of translation or of translators will be measured.

[1] https://www.merriam-webster.com/dictionary/value

One way of conceptualising such frames of reference is provided by Chesterman's (2001) discussion of translation ethics, which identifies four ethical models that foreground different values attached to translation and translators. *Norm-based ethics* emphasises the translator's role as a trusted authority, highlighting values associated with professional judgement, responsibility and expertise (Chesterman, 2001, p. 142). The *ethics of representation* foregrounds fidelity and truth, which could be linked to the *documentary value* of translation (Chesterman, 2001, p. 140). Additionally, the *ethics of communication* values the translator's function as a cross-cultural mediator, stressing localisation, cultural sensitivity and the *instrumental value* of translation in facilitating understanding (Chesterman, 2001, p. 141). Finally, the *ethics of service* underlines efficiency-related values such as time, productivity and meeting deadlines, thus focusing on the *use value* of translation (Chesterman, 2001, p. 140). Beyond these ethics-related values, translation may also be claimed to possess further values such as *epistemic value*, since it exposes readers to foreign knowledge and perspectives, or *aesthetic value*, insofar as it constitutes a creative act in the interworld between source and target languages and cultures.

Of course, the frames of reference against which the value of translation and translators is being measured are in constant flux and impacted by a range of factors. In today's phase of translation automation, it is often argued that technological change, combined with economic and discursive dynamics, may have contributed to a narrowing of the value spectrum associated with translation. Tieber and Baumgarten (2024, p. 386), for example, claim that the L&T industry has increasingly become dominated by a "spirit of instrumental thinking" and regulated by "algorithmic norms", which may undermine translators' "social and economic value as transcultural professionals". Zooming in on machine translation (MT) as one of the most influential translation technologies of the last decades (and the cradle of current GPAI LLMs), Tieber (2025, p. 227) further points out that the mere presence of MT changes the way that digitalised societies perceive the cultural practice of translation and that “the pervasive presence of MT has contributed to the reification of translation, thereby diminishing its cultural value”. The increase in automation-related pricing pressure as the main deciding factor in a translation project further emphasises this point. As one project manager in a study by Sakamoto (2021) on the value of translation in the era of automation puts it, “more and more clients are satisfied with raw MT output of sub-optimal quality ‘as long as it is not mistranslation’” (Sakamoto, 2021, p. 248). Moreover, as our subsequent analysis will show, industry discourse on accuracy increasingly focuses on the performance of technologies and efficiency, shifting towards the value of the technology itself.

From this perspective, it can be hypothesised that the ongoing automation of the L&T industry, as well as the accompanying cultural, social and economic shifts may have led to the relative devaluation of three of the four models of translation ethics discussed by Chesterman, concentrating the value of translators mostly in the *ethics of service* model—where efficiency, speed, and deliverability dominate. This may illustrate a broader dynamic of technological value destruction. As do Carmo (2025, p. 38) argues, citing Rushkoff (2016), investments in technology can paradoxically diminish value: greater productivity lowers the perceived worth of each unit of output by reducing the need for labour and by limiting avenues for quality enhancement through human input. In the translation sector, these processes are reinforced by stagnant or downward pricing, discounting tied to productivity gains, outsourcing and platformisation, and discursive narratives that rebrand services as roles while pressuring translators to evolve in lockstep with technological change (do Carmo, 2025, p. 44). Moreover, Palmer and do Carmo (2025) argue that the decreasing demand for, and therefore value of, human translation services is not only due to the ubiquity of MT and resulting pricing pressures but also due to the L&T industry's presentation of its value-added services (such as transcreation) as superior to 'simple translation'.

At the same time, the picture is unlikely to be homogeneous across the L&T industry, as different stakeholders may prioritise different values depending on their position within specific organisational and professional structures. Drawing on Freeman's (2010, p. 46) broad definition of stakeholders as actors who can affect or are affected by organisational objectives, and following Moorkens and Rocchi's (2021, p. 324) distinction between internal and external L&T stakeholders, this study considers owners, project managers, in-house translators and other employees at translation companies or departments as internal stakeholders, and clients, end users, freelance translators, language technology developers and society at large as external stakeholders. To this constellation, translation studies itself may be added as a reflexive stakeholder that actively influences and is influenced by professional discourses, norms and expectations, particularly through its impact on education and training. Against this background, the present study seeks to empirically examine how various stakeholders articulate values related to translation and translators in the current automation cycle, and how these value articulations may inform, reinforce or challenge existing approaches to translation pedagogy.

## Methodology

### Research questions and design

This study employs analytical categories for the systematic labelling and analysis of an interview corpus presented below. The analytical categories build on Chesterman's framework of translation ethics and associated values, complemented by additional value types featuring prominently in the current cycle of translation automation: *human value*, *technological value*, *efficiency* and *adaptability*. Human value is mapped to notions such as trust, expertise and validation, corresponding primarily to *norm-based ethics*, while references to culture and localisation are linked to the translator's role as a cross-cultural mediator (*ethics of communication and representation*). Efficiency-oriented values such as time pressure and deadlines are associated with the *ethics of service*, foregrounding translation's use value in increasingly automated production environments. Technological value is captured through references to the perceived performance, reliability and optimisation of translation technologies, which often position technology itself as a carrier of accuracy and efficiency. Finally, adaptability is conceptualised as a value residing at the intersection of humans and technologies, reflecting expectations that translators continually adjust their skills, roles and professional identities in response to shifting technological and organisational frames.

Based on these four categories, we set out to explore the value of translation and translators and the potential bias towards the ethics of service in our interview corpus. More specifically, we investigate the following:

1. How are human values portrayed by different L&T industry stakeholders?
2. How are technological values and the related concept of efficiency portrayed?
3. How is adaptability portrayed, and does it appear as a potential mediator between human and technological values?

***Data***

The data analysed in this study were collected as part of the LT-LiDER project (Moorkens et al., 2024), which addresses the digital competences L&T professionals require to navigate an increasingly technologised working environment. Between April and June 2024, twenty-nine stakeholders were interviewed (in 26 interviews, four per member of the consortium), representing a wide range of professional roles, institutional contexts and linguistic backgrounds. Participants were recruited leveraging the consortium contacts and included Translators and interpreters (11), Managers (10) and Academics (8) who worked in-house at language service providers (LSPs) or international organisations, as freelancers, as LSP managers, heads of research, or as academics involved in translator education. All interviews were conducted online via Zoom, recorded with participants' consent, and carried out in one of six languages—Basque, Catalan, English, French, German and Spanish. A list of interviewed participants with detailed information on profile types and all recorded videos are available on the LT-LiDER[2] project homepage. Each interview lasted approximately thirty to forty minutes and followed a shared semi-structured interview guide consisting of a script of thirteen questions addressing technology use, automation and quality practices, required skills and competences, and perceptions of the profession in the current automation cycle. Following transcription, detailed English summaries of the interviews were compiled based solely on the answers provided by the interviewees. Full transcripts were not used for the present article for homogenisation purposes, as the original interviews were conducted in six different languages, not all of which are spoken by the authors. The summaries also provided a slimmed-down version of the interviews by excluding introductions and back-and-forth exchanges between the interviewer and interviewee. They were compiled into a spreadsheet format to facilitate systematic processing and constituted the basis for the first analytical layer presented in this paper. The list of candidate terms, questions, and the annotated data can be found in our repository[3].

[2] https://lt-lider.eu/

An initial synthesis of the interview data was published in the LT-LiDER project report (Secară et al., 2025), which integrates the interview findings with an extensive review of existing surveys and industry analyses. This report documents the empirical evidence and situates it within broader industry developments. Building on this synthesis, a more fine-grained analysis of the skills and values mentioned in the interviews informed the development of a skills map[4], which organises identified skills, tools and processes across different phases of the L&T services workflow, thereby providing a comprehensive overview of competences required in a technologically mediated translation landscape.

***Data analysis***

The interview summaries were subjected to a structured preprocessing stage designed to identify linguistic markers associated with the four categories presented above. This stage began with an exploratory analysis using NotebookLM, where each summary was examined for potential lexical items and synonyms linked to the categories. Keyword prompts were combined with thesaurus-based expansion to generate an extended list of candidate terms, and NotebookLM was prompted to complete the list by including other potential candidates. This list (Figure 1) was then used to manually annotate the interview summaries and assign them to one or more analytical categories. To enhance analytical robustness and ensure coherence in the application of the analytical categories, the data were independently annotated by two researchers, who compared and reviewed their annotations. Finally, Sketch Engine was used to support the qualitative content analysis process described above.

[3] https://github.com/Morgian84-new/A-technology-oriented-mapping-of-the-language-and-translation-industry
[4] https://lt-lider.eu/mapa-de-las-tecnologias-de-la-lengua/

| Value-human | Value-tech | Efficiency | Adaptability |
|---|---|---|---|
| Driving, quality, skill/s, skilled, judgment/s, validation, language experts?, suitable?, creativity, knowledge, evaluate, evaluation, judge, (cultural) understanding, tone interpretation, non-verbal cues, human intervention, professional linguists, authorise, linguistic instincts, style, authenticity of language, unique vision, critical capacity, textual analysis, humanistic knowledge, general knowledge, intercultural mediator, human touch, refinement | accuracy | Efficient, productivity, productive, fast, speed, time, saving, optimisation, agile, automation, reduce, costs, cost-effective, workflow process, shortening time, cutting down QA time, saving time typing, preparing environment, hyperefficient, ultraefficient | train, retrain, skills, change, adaptation, adjusting, adjust, openness, learn, integrate, transfer, tech-savvy, **perspective, mindset, modifying skill set, rearrangement of skills, new skills, continuous learning, evolve/evolving, transformation, revolution, linguistic skills, technical knowledge, specialise, conform, acclimate** |

**Figure 1.** Final list of candidate terms

Following the first annotation iteration, an inter-annotator agreement (IAA) was computed, using Krippendorf's α (Krippendorff, 2019). Specifically, for each of the 13 interview questions, and overall, an IAA was computed between the two annotators based on the questions and on the labels. These IAAs should be interpreted with caution because a) multi-label annotation complicates calculation, as annotators could assign up to four labels or any combination, and b) questions differed in the number of responses, resulting in uneven totals of labelled instances per question.

***Results***

This section presents the results of the interview annotations and linguistic analysis. Figure 2 shows IAAs per question and overall (in red) for all 26 interviews. The overall IAA across all 13 questions and labels is α=0.61, slightly below Krippendorff's threshold for tentative conclusions (0.67–0.80). This reflects both topical diversity in responses and annotation variability due to the multi-label approach. Of the 13 questions, eight fall below this threshold. The lowest IAA (α=0.2661) was observed for question 2 ("How do you see the past, present and future of translation technologies? ..."), a multi-faceted question that elicited diverse responses and a balanced distribution across all four labels. Questions 1, 5, 6, 7, 8, and 13 similarly fall below the threshold, while five of the 13 questions show IAAs above the acceptable threshold, with three exceeding the 0.81–1 range considered acceptable for definite conclusions. The highest IAA (α=0.85) was obtained for question 4, “Which assignments/tasks/workflow stages are more technology-prone?”.

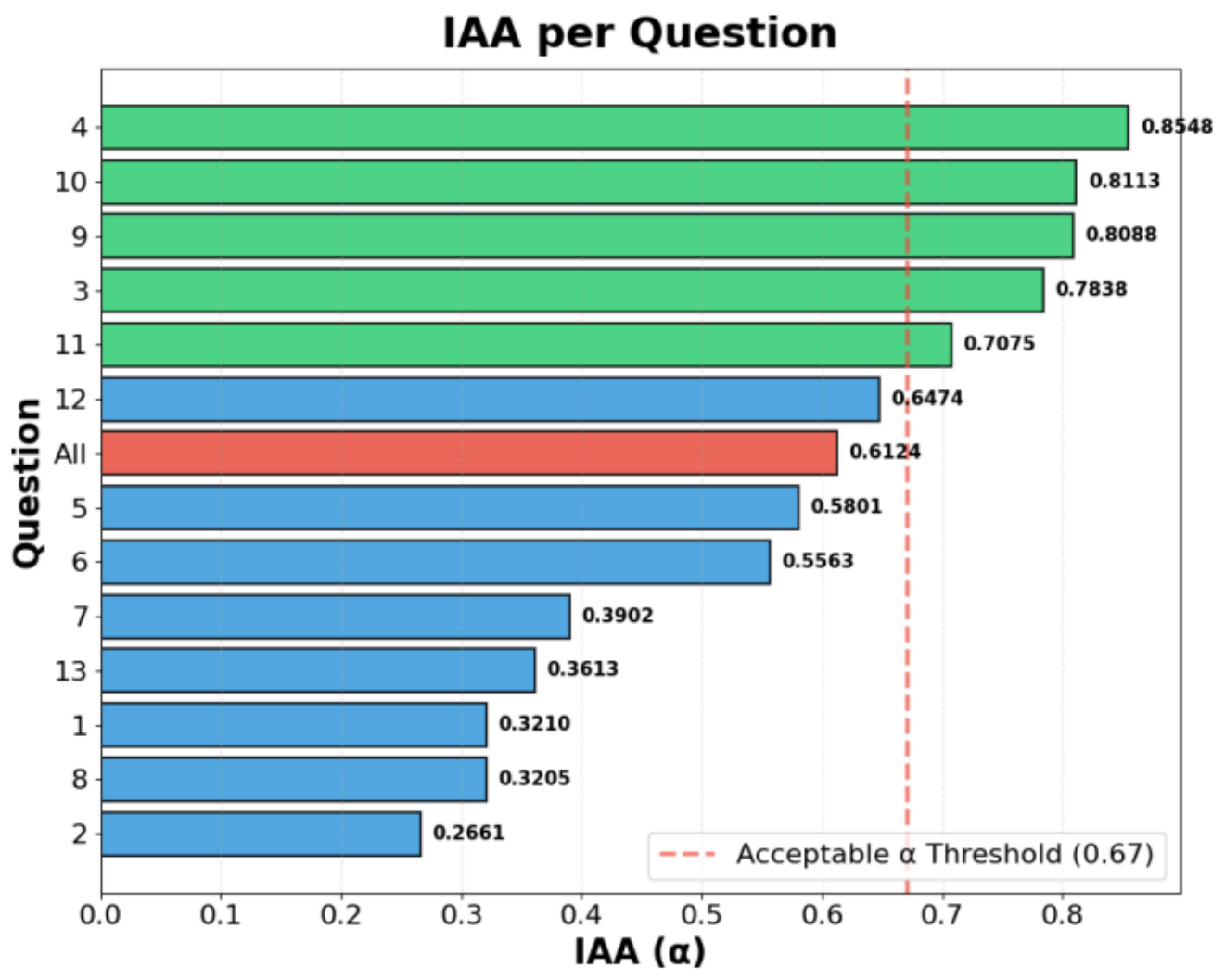

**Figure 2.** Plots for inter-annotator agreement (Krippendorff’s alpha) per question.

Figure 3 below further shows that *adaptability* is by far the most frequent label, accounting for 48% of all instances, followed by *human value* at 26%. In comparison, *technological value* and *efficiency* account for 12% and 14%, respectively. Adaptability is thus the most prominent notion across interview responses. Question 9 (“Do you see any changes in the agency and competences of translators in the future?”) exemplifies this pattern, with *adaptability* clearly dominating, followed by *human value*. Technological value appears as the most frequent label only once, in responses to question 2 discussed above. Interestingly, with α=0.78, the highest IAA computed for a single label was for *human value*. Together with this label’s occurrence, this highlights that this topic was not only discussed frequently but also with importance and clarity. Figure 3 portrays the breakdown of labels per type of stakeholder and shows that references to *adaptability* and *efficiency* are more frequent among managers, whereas translators have the highest occurrence of references to *human-value* (even though this is rather balanced across all stakeholders) as well as, surprisingly, *tech-value*.

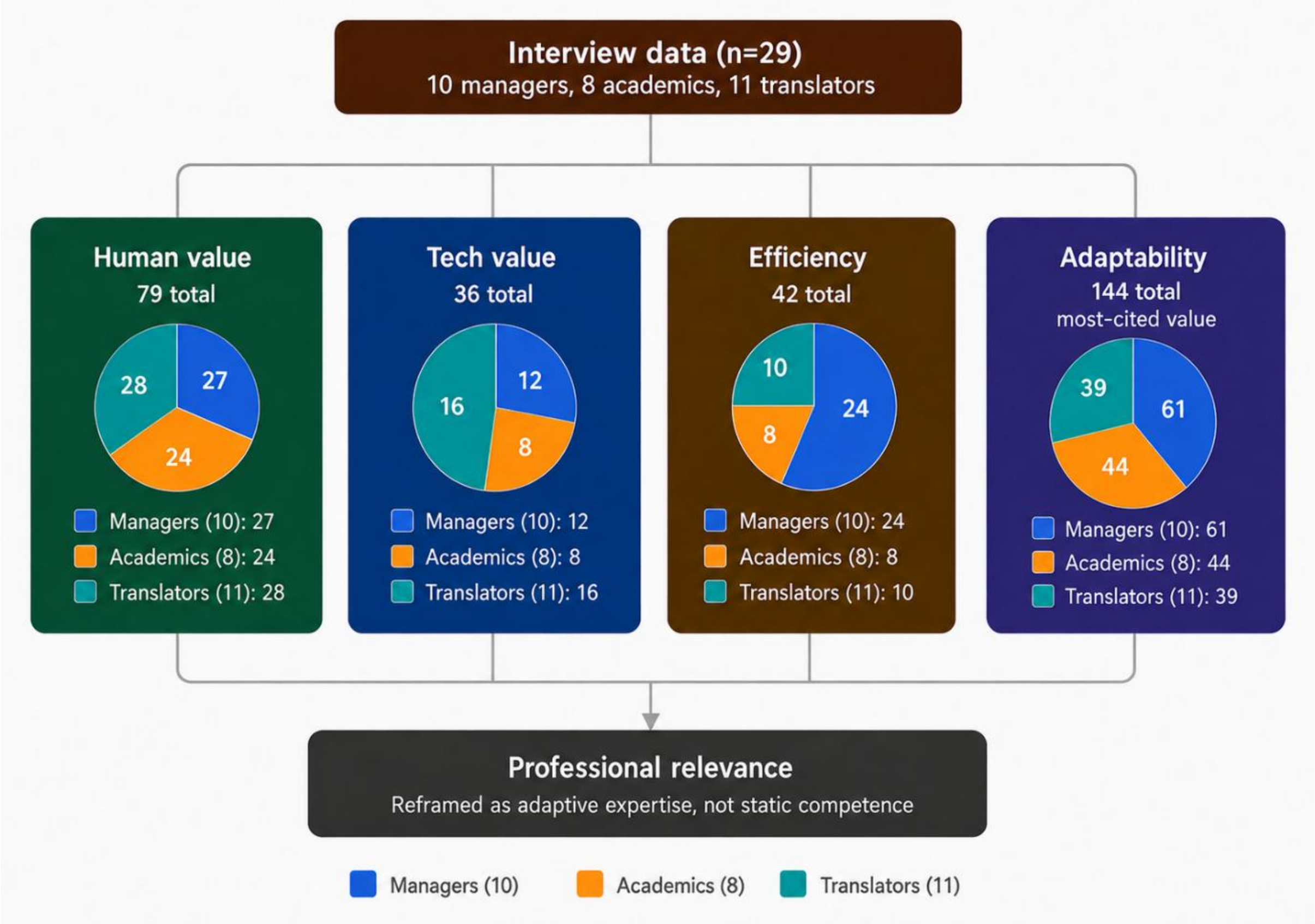

**Figure 3.** Value frequency per stakeholder profile.

In terms of our research question 1, the overall linguistic analysis foregrounds "human" as a form of value added to quality assurance, particularly in relation to accumulated professional expertise. As illustrated in Figure 4, nouns modified by *human* (65 in total, representing 0.14% of the whole corpus) cluster around 'expertise', 'review', and 'oversight', constructing human involvement as a qualitative layer within translation workflows. In contrast, technology (technological appears 27 times, 0.06% of the whole corpus) is framed in terms of 'evolution' and 'innovation' within increasingly automated environments (automatic appears 22 times, representing 0.05% of the whole corpus), including 'quality assurance' and 'correction'. This positions automation as a driver of efficiency-oriented change.

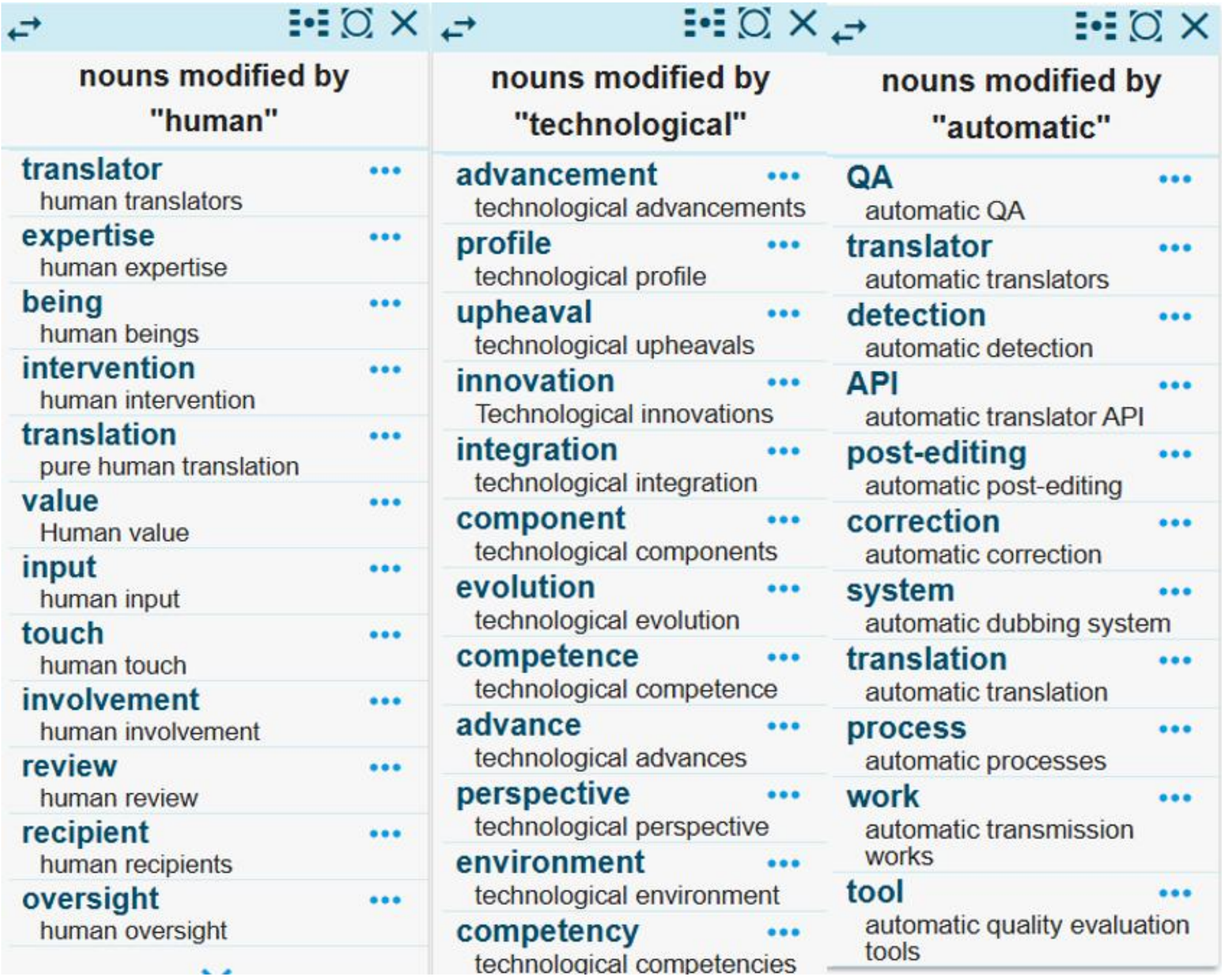

**Figure 4.** Sketch Engine word sketches for POS Adj. 'human', 'technological' and 'automatic'

This pattern evidences a strong trend in our interview data, where respondents repeatedly articulate *human value* (79 occurrences) as a stabilising and evaluative factor within technology-mediated production contexts. One interviewee explicitly states that "the ability to translate well has been crucial in assessing any translation, either by humans or machines" (AR3, freelance translator), positioning human competence as the benchmark against which both human and machine output are evaluated. This formulation implicitly establishes human expertise as a meta-level criterion, extending beyond manual translation to encompass the assessment of automated systems themselves. *Human value* is thus not confined to production alone but also resides in evaluative and supervisory practices grounded in accumulated professional experience. Here, quality is conceptualised as something that must be actively judged rather than automatically assumed.

In discussing the limitations of MT, one respondent notes that "for certain languages such as Romanian, MT is not really an option" (K3, freelance translator), pointing to gaps in data coverage and performance that necessitate continued human intervention. Here, *human value* is conceptualised as a form of situated knowledge compensating for uneven technological development. This aligns closely with the word sketch results, where *human* modifies nouns such as ‘expertise’ and ‘intervention’, reinforcing the idea that human involvement is invoked particularly in contexts where quality is at stake.

The interviews also reveal that *human value* is closely tied to accountability and responsibility towards recipients. One respondent highlights that poorly translated content would cause users to “lose trust in that translation” (AU3, freelance translator), suggesting that human oversight functions as a guarantor of credibility and reliability. This interpretation is supported by the presence of human review in the word sketch data, which frames translation as a relational practice oriented towards end users rather than a purely technical process. In this sense, human value is associated with *norm-based ethics*, the *ethics of communication,* and the *ethics of service*, including responsibility for communicative outcomes and the maintenance of trust.

Importantly, human values do not stand in opposition to technology. Instead, they are articulated as complementary to technological systems viewed in terms of innovation and evolution. While technology is often framed as efficient and optimised, human involvement is repeatedly highlighted as necessary to ensure that these systems function reliably in real-world conditions. Human oversight, review, and expertise thus operate as quality-assurance layers in automated workflows. Our data analysis suggests that human value in contemporary translation discourse is constructed as an enabling and legitimising factor. It contributes interpretive judgment, accountability, and contextual awareness to increasingly automated production environments, foregrounding translation’s use value while maintaining a combined norm-based ethics of service oriented towards quality, trust, and professional responsibility.

Building on RQ1 and RQ2, our findings support Palmer and do Carmo’s (2025) claim that technological value/efficiency and text customisation (human value) constitute the two primary differentiators between MT and human translation, while also offering a more nuanced ethical interpretation derived from our interview data. Stakeholders consistently portray technology as a means to achieve the efficiency required in fast-paced, high-output environments, with speed reconceptualised not as an individual ability but as the result of streamlined tools and processes: “as speed becomes the main concern, the question is not how fast you can do it, but which processes and tools you can apply” (O5, IT manager). As per Figure 5, while only people are *adaptable* (2 in total, representing 0.004% of the whole corpus), mostly working with technology (*TMSs* and *manner*) tends to be discussed as *efficient* (8 times, 0.02 % of the whole corpus). Automation is explicitly associated with reducing time pressure and turnaround constraints and is frequently evaluated more favourably than human translation when efficiency alone is at stake, aligning closely with Chesterman’s *ethics of service* and foregrounding translation’s use value. At the same time, this efficiency is rarely presented as an end in itself. Rather, respondents implicitly describe efficiency as a prerequisite that saves time and frees cognitive resources for translators to engage in less efficient, but more value-adding, forms of work such as text adaptation, contextualisation, and adequacy-focused customisation. In this sense, technological efficiency functions as an infrastructural condition that enables, rather than replaces, the *ethics of communication*, allowing translators to focus on tailoring texts to specific briefs, audiences, and communicative goals.

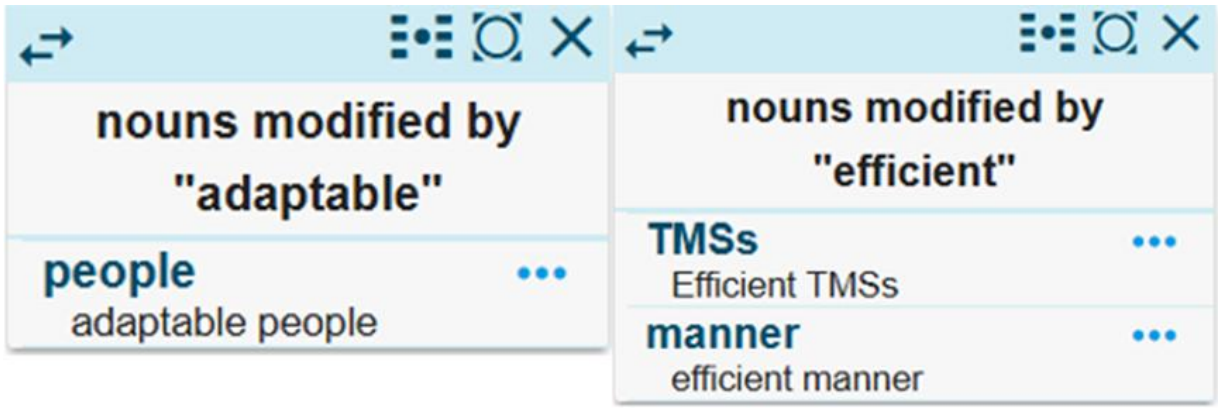

**Figure 5.** Sketch Engine word sketches for POS Adj. ‘adaptable and ’efficient’

The data suggest a complementary relationship between the *ethics of service* and the *ethics of communication*, rather than a simple hierarchy in which efficiency outranks communicative or cultural concerns. While the primacy of the *ethics of service* is supported insofar as efficiency and deliverability are treated as non-negotiable baseline requirements, automation is frequently evaluated positively for its capacity to absorb repetitive or time-consuming tasks. Human translation, by contrast, is valued for its ability to interpret intent, manage ambiguity, and adapt content in ways that exceed the capabilities of current MT systems and LLMs. This division of labour reinforces the idea that efficiency gains derived from technology are instrumental in sustaining human-led text customisation, which remains comparatively resource-intensive and resistant to full automation[5]. Rather than signalling the primacy of the *ethics of service* alone, the interviews point to its role as a necessary support structure for the ethics of communication in a highly automated L&T industry. Efficiency-oriented technological practices ensure deliverability and economic viability, while human translators are expected to contribute their expertise to guarantee communicative effectiveness and contextual appropriateness.

The verb patterns associated with *translator* and *technology* in Figure 4 further reinforce adaptability as a central value impacting the contemporary professional profile, and, as per Figure 6, to avoid the constant ‘pressure’ of being ‘replaced', translators need to ‘train’ and ‘prepare’. The lemma *adapt* appears 27 times (0.06% of the corpus) and the top three collocations associated with it are ‘need to’, ‘have/had to’, ‘how to’. [67] who must train and prepare in order to remain professionally viable, often in response to sustained pressure and concerns about being replaced by technology. One interviewee explicitly notes that translators need to actively develop technological skills to cope with change (W28, academic), while another refers to the profession as being under constant pressure to keep up with technology (K23, MT manager). At the same time, technology is predominantly constructed through verbs such as ‘help’, ‘improve’, and ‘evolve’, reflecting interview narratives in which technology is conceptualised as reconfiguring workflows instead of fully substituting human expertise. These verbal choices, particularly ‘help’ and ‘improve’, were relatively balanced throughout stakeholder groups, while ‘evolving’ was more frequently used by managers. This highlights adaptability as a relational competence resulting from the interaction between translators and technologies: while translators are expected to continuously adjust their skills and professional identities, technology is framed as something to be integrated, leveraged, and understood, positioning adaptability as a key value mediating between human agency and technological change.

---

[5] This is conceptualised both as a current and a future trend, as our interviewees’ views often looked at future prospects (“will” is frequent, as shown in the above word cloud).

[7] A link between human agency and value in a translation context is also established by Massey (2025, p. 11), who observes that “the concept of agency has become a yardstick by which the value of professional human translation can be judged”.

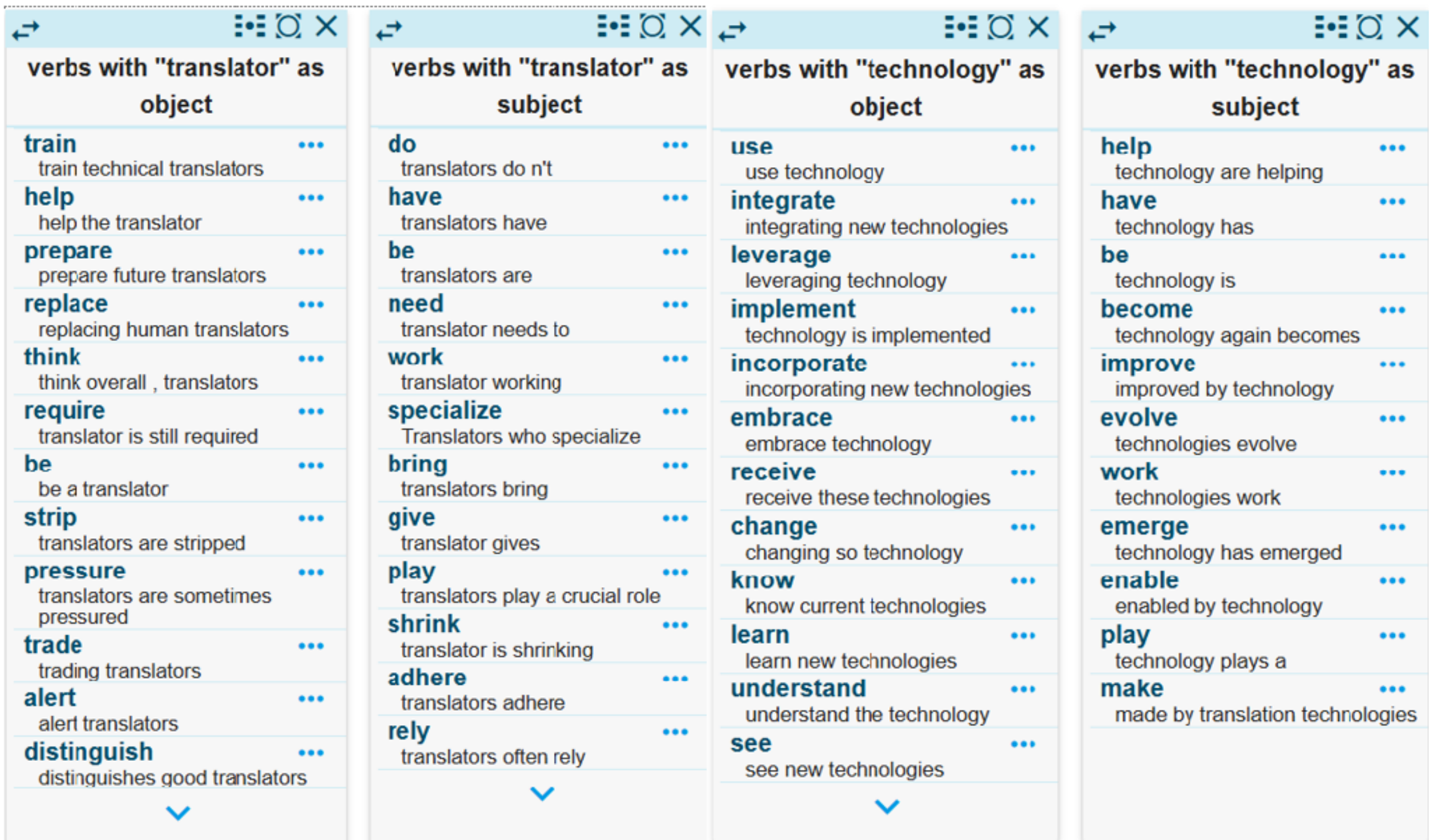

**Figure 6.** Sketch Engine word sketches for POS N. ‘translator’ and ‘technology'

As the LT-LiDER Language Technology Map illustrates, the skills identified by respondents are not limited to tool operation (considered as the basic/standard manipulation of a tool for a given task). They encompass higher-level digital literacy, such as understanding the capabilities and limitations of MT and LLMs, effective prompt design, and the strategic use of automation to streamline QA processes. These competences enable translators to work flexibly across different stages of the workflow and respond to shifting project requirements. Thus, adaptability is understood not as reactive compliance with technological change, but as an active capacity to influence how technology is used to achieve communicative goals.

This perspective aligns with the view of translation as inherently adaptable. As do Carmo and Moorkens argue, translation has long been characterised as a “secondary, efficient, and adaptable communication process” aimed at enhancing an original communicative act (2022, p. 20). The interview data suggest that this adaptability now manifests not only in text production but also in service configuration. Translators add value when they can support both text customisation (cultural adaptation to purpose) and service customisation (leveraging client knowledge, domain expertise, and workflow optimisation) (Palmer and do Carmo 2025, p.17).

From this perspective, adaptability mediates between efficiency-oriented technological practices and the ethics of communication. By offloading routine tasks to automation, adaptable translators can invest time and expertise in interpretive, contextual, and relational aspects of translation work. The skills identified in the LT-LiDER Language Technology Map support Massey's (2025, p. 11) notion of consolidating human–agent interaction competence, where professionals are not displaced by generative AI but are required to interact with it critically, reflexively, and strategically. Adaptability thus becomes essential for sustaining professional relevance and ethical agency in an increasingly automated L&T industry.

**Discussion and pedagogical implications**

This study examined how stakeholders articulate values around translators and translation amid accelerating automation. Our interviewees highlight both adaptability and technology-based efficiency as fundamental values in the translation industry, which aligns with previous research (do Carmo and Moorkens 2022; Palmer and do Carmo 2025). According to do Carmo and Moorkens, an efficient translation process uses "the best available methods and technologies to produce a rapid outcome", and an adaptable one has "a mouldable shape, including creativity and [can adjust] to different requirements" (2022, p. 20). Similarly, Palmer and do Carmo observe that efficiency is "especially close to the heart of the productivity-focused language industry" (2025, p. 16). Building on these conceptualisations, our study examined both adaptability and the efficiency dimension of technological value as well as how human and technological value are negotiated in contemporary constructions of the translator's role. The interview data reflect value being attributed both to the skilled and qualitative human, and to the speed, scalability, and ease afforded by technology.

Overall, the linguistic analysis and the LT-LiDER Language Technology Map reveal that values in the AI-automated L&T industry have clearly shifted. Human values such as expertise, judgment, agency and accountability remain central, but they are increasingly articulated in direct relation to technological systems and situated within efficiency-oriented production environments. Meanwhile, adaptability is conceptualised as a key mediating value, linking human expertise to technological change and reframing professional relevance as contingent on continuous skill and competence development.

A prominent finding is the growing emphasis on the alignment of technological value with the *ethics of service*. Across the interviews, efficiency, speed, and deliverability are treated as baseline requirements. Automation is viewed positively when it relieves time pressure, speeds up turnaround, or takes over routine tasks—particularly in QA and pre-production stages. This supports Palmer and do Carmo's (2025, 22) claim that efficiency constitutes a primary differentiator between MT and human translation, and that technological value increasingly structures how translation work is organised and assessed. However, this development does not straightforwardly devalue human expertise. Instead, this perspective emerges most prominently in managers' discourse, where efficiency is considerably more salient than among translators or academics (24 mentions, compared with 10 and 8, respectively). Managers tend to view these efficiency gains as enabling translators to focus on text customisation, contextualisation, and communicative effectiveness—activities that still rely on human judgment and the *ethics of communication*.

Professional competence is, however, being redefined. Traditional views of "the good translator" prioritised linguistic mastery and cultural mediation as self-contained human attributes (Dam et al., 2025, p. 2). By contrast, our interviewees highlight the importance of acquiring new technological and interactional skills. Translators are expected to understand, integrate, and critically supervise technologies such as CAT tools, MT, and LLM-based systems. This contrasts with the findings of Dam et al. (2025), whose interviewees described a more traditional self-image in which technology remains a background resource and not a core competence. A plausible explanation for this divergence lies in timing: Dam et al.'s interviews were conducted in 2022 only with translators, prior to the rapid uptake and discursive dominance of generative AI, whereas the LT-LiDER interviews were conducted in 2024, when LLMs had become highly visible, contested, and consequential actors in the industry. Moreover, out of 26 interviewees we only had 11 translators and interpreters. This suggests that value constructions in the L&T sector are both stakeholder-dependent and highly sensitive to rapid technological cycles.

In an environment of rapid change and uncertainty, translators are expected to continuously update their skills and competences to remain viable, highlighting the fact that technological competence transcends 'technological proficiency' as a stand-alone category. Thus, adaptability is a core professional value, closely tied to employability, professional identity, and ethical agency. This aligns closely with the LT-LiDER Language Technology Map, which documents a broad and expanding range of competences spanning the entire translation workflow, from prompt design and MT quality estimation to data curation and automated QA. The map illustrates that adaptability is not a form of ad hoc reskilling, but rather a structured and cumulative competence integrating linguistic, technical, and strategic dimensions.

***Pedagogical implications***

These findings have clear implications for translation pedagogy. They suggest that the times are long gone when technology could be treated as a discrete or auxiliary component of translator training. Instead, it must be embedded throughout curricula and explicitly linked to ethical decision-making, quality assessment, and communicative goals. In the same way that the increasing interdependence between humans and modern AI technologies that is intended to benefit all stakeholders is often framed within the *human-centered AI* approach (cf. Jiménez-Crespo, 2025)—where humans retain high levels of autonomy and control in highly automated workflows—translation pedagogy should commit to “human-centered pedagogies in the age of generative AI” (Ayvazyan, 2025). This requires a multidimensional approach at least along the four axes of *human value* (‘traditional’ translation competences embedded in an ethical framework as discussed in this paper), *technological value* (technological skills as outlined in the LT-LiDER Language Technology Map and coordinated through higher-level digital and AI literacy as addressed by the overall LT-LiDER project), *efficiency* (identification of automation potentials and required added value of human experts, again underpinned by an ethical framework) and *adaptability*. This last aspect was discussed prominently in the interviews forming the empirical basis of this paper and therefore deserves a brief separate discussion. As our analysis showed, adaptability resides at the intersection of human expertise and (constant and accelerating) technological change and thus calls into question any static and final notions of human competence. An interesting concept in this regard is *adaptive expertise*, which was introduced into translation didactics by Angelone (2022). Adaptive expertise "involves optimal performance in contexts where tasks are more ambiguous and where they call for novelty in problem-solving" (Angelone, 2022, p. 65). Specifically, this means foregrounding learning-to-learn skills, critical engagement with tools, and reflective practice, rather than focusing solely on mastering specific technologies that may rapidly become obsolete. Strategically, fostering adaptive expertise in translator training requires orienting training towards professional realities, both in terms of career paths and actual working practices and breaking down intra- and inter-curriculum silos "so that students can see the bigger picture of the language industry and the many moving parts it entails" (Angelone, 2022, p. 71). This could be realised, e.g., within multi-stakeholder frameworks such as the European Commission's European Master's in Translation (EMT) network, which brings together translator training programmes from EU member states as well as representatives from the Language Industry Expert Group (LIND) to discuss academia-industry alignment. Multidimensional translation pedagogy, as outlined here, is also addressed by the LT-LiDER project, under the auspices of which the interview data analysed in this paper were collected. In addition to the LT-LiDER Language Technology Map, the project provides the didactically oriented NMT platform *ProMut*[8], conceived to learn how to create, manage and evaluate NMT engines. In addition to these more technology-oriented resources, LT-LiDER is currently developing a self-assessment questionnaire[9] to assist interested parties in self-assessing their translation-related digital literacy, as well as a range of training capsules[10], which will allow translation educators and students solve specific automation-related tasks in specific translation scenarios. The upcoming LT-

[8] https://lt-lider.eu/promut/
[9] https://lt-lider.eu/cuestionario-de-autoevaluacion/
[10] https://lt-lider.eu/pildoras-formativas/

LiDER textbook[11] will provide a higher-level conceptual framework for the project. Taken together, the scenario-based activities and reflective materials provided by LT-LiDER operationalise adaptability, digital and AI literacy, and human–machine interaction competence, helping align academic training with evolving professional expectations.

## Conclusion

This study shows that value in the AI-automated language and translation industry is not displaced from humans to technology but reconfigured through their interdependence. Human added value is increasingly conceptualised as a critical engagement with technological systems, including oversight, judgement, and contextual awareness, while technological value gains legitimacy through integration of technologies in human-supervised workflows that ensure quality and trust. Drawing on linguistic analysis of interview data, and the LT-LiDER Language Technology Map, the study demonstrates that efficiency-oriented technological values aligned with the ethics of service have become baseline conditions structuring professional translation practice. These developments, however, do not erode human value; instead, expertise, accountability, and evaluative competence are repositioned as higher-order functions within automated environments. A key contribution of the paper is the identification of adaptability as a mediating value linking efficiency-driven technological practices with enduring ethical commitments to communication. Adaptability reframes professional relevance as a dynamic capacity grounded in adaptive expertise. The findings have implications for translation pedagogy, highlighting the need to integrate technology, ethics, and adaptability in ways that align industry trends with the communicative and interpretive core of translation practice.

## Acknowledgements

LT-LiDER has been funded with support from the European Commission. This publication reflects the views only of the authors, and the Commission cannot be held responsible for any use which may be made of the information contained therein.

## Declaration of interest

No potential conflict of interest was reported by the authors.

[11] https://lt-lider.eu/el-libro/